\documentclass[10pt,final,journal,twocolumn]{IEEEtran}
\usepackage{lineno}
\usepackage{amssymb, amsmath}
\usepackage{amsmath,amsfonts,amssymb,amsthm,epsfig,epstopdf,url,array}
\usepackage{graphicx}
\usepackage[justification=centering]{caption}
\usepackage{multirow}
\usepackage{hyperref}
\usepackage{enumerate}
\usepackage{subfig}
\usepackage{color}
\usepackage{xcolor}
\usepackage{float}

\usepackage[flushleft]{threeparttable} 
\usepackage{booktabs,caption}

\usepackage{pgf}
\usepackage{tikz}
\usetikzlibrary{arrows,automata}

\definecolor{azulpeluso}{RGB}{143,184,229}

\renewcommand{\qedsymbol}{$\blacksquare$}

\ifCLASSOPTIONcompsoc
  \usepackage[nocompress]{cite}
\else
  \usepackage{cite}
\fi

\ifCLASSINFOpdf
\else
\fi

\begin{document}
%

\title{Short-term Cognitive Networks, Flexible \\Reasoning and Nonsynaptic Learning}

\author{Gonzalo~N\'apoles,~Frank Vanhoenshoven,~Koen~Vanhoof
\IEEEcompsocitemizethanks{\IEEEcompsocthanksitem G. N\'apoles, Frank Vanhoenshoven and K. Vanhoof are with the Faculty of Business Economics, Hasselt Universiteit, Belgium. \protect\\
E-mail: gonzalo.napoles@uhasselt.be

}
}

\IEEEtitleabstractindextext{%
\begin{abstract}
While the machine learning literature dedicated to fully automated reasoning algorithms is abundant, the number of methods enabling the inference process on the basis of previously defined knowledge structures is scanter. Fuzzy Cognitive Maps (FCMs) are neural networks that can be exploited towards this goal because of their flexibility to handle external knowledge. However, FCMs suffer from a number of issues that range from the limited prediction horizon to the absence of theoretically sound learning algorithms able to produce accurate predictions. In this paper, we propose a neural network system named \emph{Short-term Cognitive Networks} that tackle some of these limitations. In our model weights are not constricted and may have a causal nature or not. As a second contribution, we present a nonsynaptic learning algorithm to improve the network performance without modifying the previously defined weights. Moreover, we derive a stop condition to prevent the learning algorithm from iterating without decreasing the simulation error.
\end{abstract}

}

\maketitle

\IEEEdisplaynontitleabstractindextext

%

\section{Introduction}
\label{sec:introduction}

Fuzzy Cognitive Maps (FCMs) \cite{Kosko1986} \cite{Kosko1988} continue to grow in popularity mainly because of their potential to deal with expert knowledge. In these recurrent neural networks, neurons have a specific meaning for the problem under investigation, whereas edges denote causal relationships \cite{Felix2017a}. In a simulation context, the reasoning is devoted to computing the system state attached to an initial vector, which is regularly provided by the expert. This is equivalent to computing the activation value of each decision neuron from non-decision ones.

Likewise, fuzzy cognitive mapping has been used to design more complex machine learning solutions. The development of forecasting models for univariate / multivariate time series \cite{Pedrycz2014b} \cite{Pedrycz2015a} \cite{Froelich2017a} \cite{Salmeron2016a} and granular cognitive classifiers \cite{Napoles2016c} \cite{Napoles2017c} are examples that illustrate the potential attached to this technique. Even the \emph{Neurocomputing} journal recently dedicated a special issue \cite{Froelich2016a} to relevant theoretical advances in this field. But to what extent some contributions reported in the literature can be considered authentic FCM solutions (as originally defined by Kosko in \cite{Kosko1986} \cite{Kosko1988}) may be questionable. For example, would it be correct to claim that an FCM model adjusted using a data-driven heuristic learning algorithm properly reflects real-world causalities between any two concepts?

Generally speaking, FCMs are far from being a theoretically robust simulation technique. The rather limited prediction horizon of neurons, the absense of an accurate learning algorithm and the ambiguous semantics of fixed points are examples of shortcomings identified in this model. Carvalho \cite{Carvalho2010} discussed some of these problems, but for some reason the current FCM research remains inside the box. While some of the above-mentioned drawbacks seem perfectly solvable, the fact is that traditional FCMs face theoretical barriers difficult to surmount. For example, in \cite{Napoles2018a} the authors introduced the contraction theorem for FCMs which states that the activation space of a sigmoid neuron may shrink infinitely, without any guarantee of reaching an equilibrium point. Even if the system converges, there is no guarantee that the discovered equilibrium attractor leads to the lowest simulation error.
   
In spite of the above-mentioned issues, the cognitive mapping principle stands as a powerful approach to perform simulations on the basis of previously defined knowledge structures. In our research, we refer to this valuable characteristic as the \emph{flexible reasoning}. Not many machine learning techniques allow to directly embed knowledge into their reasoning process. It is worth mentioning that in the flexible reasoning paradigm the knowledge structures can be defined by experts or computed by other algorithms. However, designing such a highly flexible, sound simulation model under the umbrella of traditional FCMs may become (unnecessarily) difficult. For example, what if the experts have a clear picture of how the variables correlate, but they are unable to claim the existence of causal relationships between them?

This paper brings to life the following contributions. Firstly, we discuss some major problems affecting FCM-based models which motivated this research. Secondly, we introduce a neural system referred to as \emph{Short-term Cognitive Networks}, which allow reasoning on the basis of previously defined knowledge structures. In our research such structures refer to the weight matrix that defines the interaction among system variables. Thirdly, we propose a nonsynaptic learning algorithm to reduce the global simulation error without modifying the weight matrix. This gradient-based algorithm regulates the excitation degree of each neuron in each iteration. As a last contribution, we analytically derive a stop condition to prevent the learning algorithm from iterating without decreasing the global error, thus notably increasing its efficiency.
 
The rest of this paper is organized as follows. Section \ref{sec:fcm} provides a concise background on traditional FCMs, whereas Section \ref{sec:motivation} discusses the key motivations behind our proposal. Section \ref{sec:stcn} presents our flexible reasoning model baptized as \emph{Short-term Cognitive Networks}, while Section \ref{sec:learning} is dedicated to the nonsynaptic learning method. Section \ref{sec:sim} introduces the numerical simulations, and Section \ref{sec:conclusions} outlines the concluding remarks and future research avenues.

\section{Fuzzy Cognitive Mapping}
\label{sec:fcm}

\begin{figure}[!ht]
\begin{center}
\begin{tikzpicture}[->,>=stealth',shorten >=1pt,auto,node distance=4cm,
                    semithick]
  \tikzstyle{every state}=[fill=azulpeluso,draw=azulpeluso,text=black]

  \node[state] 		(A)              		{Predator};
  \node[state]      (B) [right of=A] 		{Prey};
  \node[state]      (C)	[below right of=B] 	{Grass};

  \path (A) edge [bend left]  node {$-1.0$} (B)
        (B) edge [bend left]  node {$1.0$} (A)
        (B) edge [bend left]  node {$-1.0$} (C)
        (C) edge [bend left]  node {$1.0$} (B);
        
\end{tikzpicture}
\caption{FCM-based system with three concepts modeling the interrelations in a simplified food chain.}
\label{fig:fcm-example}
\end{center}
\end{figure}
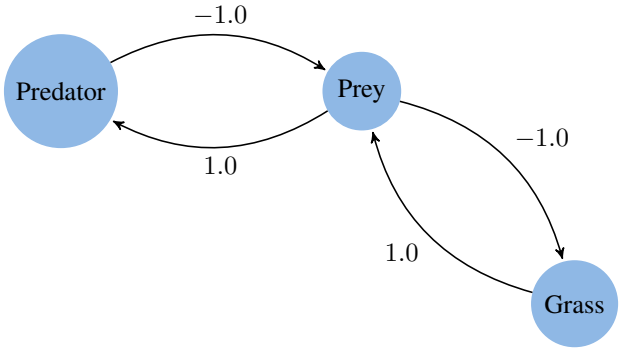

In a nutshell, an FCM denotes a complex network of causal relationships between abstract concepts. Figure \ref{fig:fcm-example} portrays an FCM describing the complex interrelations in a simplified food chain. The model is intuitive and fairly self-explanatory. It can be seen that predators thrive when there is a lot of prey, whereas the prey animals themselves are under pressure from the predators but can be boosted by the presence of grass. Given a set of predators, prey and grass, the FCM can evolve in several \emph{iterations}. It is likely that after a sufficient number of iterations, the network will eventually find a stable state in which the entire system reaches a balance.

In its mathematical form, an FCM can be defined by a 4-tuple $\langle \mathcal{C}, \mathcal{W}, \mathcal{A}, f(\cdot) \rangle$, where $\mathcal{C} = \{C_1, \dots C_M\}$ comprises the variables describing the physical system, typically visualized as nodes in the network while $\mathcal{W}$ denotes an $M \times M$ weight matrix where $w_{ij} \in [-1,1]$ encodes a causal relation of $C_i$ upon $C_j$. In the graph, each weight is visualized as a labeled edge between the corresponding concepts. The interpretation of causal weights is depicted as follows: 

\begin{itemize}
  \item $w_{ij} > 0$: If activated, $C_i$ will \emph{excite} $C_j$. More explicitly, higher (lower) activation values of $C_i$ in the current iteration will lead to higher (lower) activation values of $C_j$ in the following iteration;
  \item $w_{ij} < 0$: If activated, $A_i$ will \emph{inhibit} $C_j$. More explicitly, lower (higher) activation values of $C_i$ in the current iteration will lead to higher (lower) activation values of $C_j$ in the following iteration.
  \item $w_{ij}=0$: $C_{i}$ will not influence $C_{j}$. In the graphical model, this relation is usually indicated by the absence of an edge between the two concepts.
\end{itemize}

These statements are different from common phrases like \emph{an increase (decrease) in one will cause an increase (decrease) in the other}. The systematic misinterpretation of causal weights has been widely discussed by Carvalho \cite{Carvalho2010}. In this paper, we have chosen a definition that is more in line with the inference rule attached to the neural system. 

The function $\mathcal{A}: \mathcal{C} \times \mathbb{N} \rightarrow A_i^{(t)}$ comprises the activation value of the $C_i$ concept in the $t$th iteration step. The initial activation value $A_i^{(0)}$ can be either determined by the expert at the beginning of the simulation, or computed from historical data. During the reasoning process, such values are updated using the inference rule depicted in Equation \ref{eq:kosko}. In reference to the previous paragraph, it can be seen that $A_i^{(t+1)}$ is defined by the activation values of its connected concepts in the previous iteration, not by any increase or decrease. On the other hand, this reasoning rule treats the concepts as standard McCulloch-Pitts neurons \cite{McCulloch1988} in a causal graph, which is why FCMs are considered (recurrent) neural systems.

\begin{equation}
\label{eq:kosko}
	A_i^{(t+1)}=f \left(\sum_{j=1}^M w_{ji} A_j^{(t)} \right), i \neq j
\end{equation}

A pivotal aspect of any neural model refers to the transfer function $f: \mathbb{R} \rightarrow I$ used to maintain the incoming evidence within the activation space $I$, often $I=[0,1]$ or $I=[-1,1]$. Equation \ref{eq:sigmoid} displays the sigmoid function, a common choice in fuzzy cognitive modeling because of its ability to represent both qualitative and quantitative scenarios. This function involves a parameter $\lambda>0$ that controls the function slope. Recent studies \cite{Napoles2016a} \cite{Napoles2016b} have shown that this parameter may have a key impact on the FCM convergence.

\begin{equation}
\label{eq:sigmoid}
	f(x) = \frac{1}{1+e^{-\lambda x}}
\end{equation}

For a sufficient number of iterations $T$, the recurrent system can be attracted to one of the following states:

\begin{itemize}
  \item \textbf{Fixed-point} $(\exists t_{\alpha} \in \{1,2,\dots,(T-1)\} : A^{(t+1)}=A^{(t)}, \forall t \geq t_{\alpha})$: the map produces the same output after the cycle $t_{\alpha}$, so $A^{(t_{\alpha})}=A^{(t_{\alpha}+1)}=A^{(t_{\alpha}+2)}=\dots=A^{(T)}$.
  \item \textbf{Limit cycle} $(\exists t_{\alpha},P \in \{1,2,\dots,(T-1)\} : A^{(t+P)}=A^{(t)}, \forall t \geq t_{\alpha})$: the map produces the same output periodically after the cycle $t_{\alpha}$, so $A^{(t_{\alpha})}=A^{(t_{\alpha}+P)}=A^{(t_{\alpha}+2P)}=\dots=A^{(t_{\alpha}+jP)}$ where $t_{\alpha}+jP \leq T$, such that $j \in \{1,2,\dots,(T-1)\}$.
  \item \textbf{Chaos}: the map continues to produce different state vectors for successive cycles.
\end{itemize}

There does not seem to be an academic agreement on the interpretation of an iteration in an FCM. Typically, the meaning of an iteration is ignored when modeling simulation or pattern classification scenarios. In these cases, it is desired that the system converges to a fixed point in order to draw conclusions. In time series forecasting, however, iterations are typically equated with the time span between two discrete observations in a dataset. Fixed-point attractors are less desirable in time series forecasting since such equilibrium points suppress the model's ability to forecast fluctuations.  

\section{Criticism and Motivation}
\label{sec:motivation}

For many years we have presented FCMs as an almost infallible technique for modeling and simulating complex systems. However, as already mentioned, FCM-based models exhibit serious drawbacks that may affect the simulation outcomes. In this section, we discuss three of such issues: (1) the causation fallacy behind existing learning algorithms, (2) the restrictions imposed by causal weights, and (3) the implications of unique attractors in simulation scenarios.

According to Kosko \cite{Kosko1988}, each arrow in an FCM represents a causal relationship that is described by a signed and bounded weight. Observe that the causality assumption is pivotal here, otherwise we are in presence of traditional associative neural networks. However, the most successful learning algorithms --- which are moslty based on heuristic search methods --- cannot produce authentic causal structures since they basically perform like black boxes. This implies that we cannot claim any kind of causation unless we establish constraints beforehand, which must be defined by the experts. As an alternative, we can employ Hebbian-like algorithms which attempt to compute a model with minimal deviation from a predefined initial matrix. Although those methods could be useful in control problems, their poor generalization capability \cite{Felix2017a} makes them unsuitable for more challenging prediction scenarios. 

Another issue with the fuzzy cognitive model is that causal weights are confined to the $[-1,1]$ interval. However, it can be verified that this constraint greatly reduces the prediction horizon of FCM-based models.

\emph{Proof}. Let $C_i$ be a sigmoid neuron, then its activation value in the $t$th iteration can be expressed as:

\begin{equation*}
\label{eq:act}
A_{i}^{(t)}(k)=\frac{1}{1+e^{-\lambda \left( \sum_{j=1}^M w_{ji} A_j^{(t-1)}(k) \right)}}, \forall k
\end{equation*}

\noindent where $k$ denotes the index of the activation vector used to start the recurrent reasoning process.

Since $-1 \leq w_{ji} \leq 1$ and $0 \leq A_{j}^{(t-1)}(k) \leq 1$, then

\begin{equation*}
\label{eq2}
\text{min}(C_i) \leq \sum_{j=1}^M w_{ji} A_j^{(t-1)}(k) \leq \text{max}(C_i), \forall k
\end{equation*}

\noindent where 
\begin{equation*}
\label{eq:min}
\text{min}(C_i) = \sum_{j=1}^{M} \frac{w_{ji}(1-\text{sig}(w_{ji}))}{2}
\end{equation*}

\begin{equation*}
\label{eq:max}
\text{max}(C_i) = \sum_{j=1}^{M} \frac{w_{ji}(1+\text{sig}(w_{ji}))}{2}.
\end{equation*}

Moreover, since the sigmoid function $f(x)$ is monotonically non-decreasing, we can confidently state that:

\begin{equation*}
\label{eq:interval}
\frac{1}{1+e^{-\lambda( \text{min}(C_i) )}} \leq A_{i}^{(t)}(k) \leq \frac{1}{1+e^{-\lambda( \text{max}(C_i) )}},\forall k.
\end{equation*}

Observe that the neuron will never reach values outside this activation interval, which reduces its prediction horizon. For example, let us assume that the event $C_i$ is caused by $C_1$ and $C_2$ such that $w_{1i}=w_{2i}=-1$ and $\lambda=1$, then the minimal value that $C_i$ can produce is 0.1192, regardless of the input to $C_1$ and $C_2$, or the number of iterations. \qedsymbol

It is fair to mention that we can attack this problem from different angles. For example, if we increase the lambda value to 5 in the previous example, then the activation bounds will be expanded. However, the sigmoid function will resemble the binary activator, and again we are reducing the prediction horizon. Perhaps the easiest approach to increase the prediction ability of these models is to assume that $w_{ji} \in {\Bbb R}$ which could comprise a causal meaning or not.

The last issue refers to the FCM convergence to a unique fixed point. The existence and uniqueness of the fixed-point attractor on FCM-based models is a complex problem that has been rigorously studied in the literature \cite{Boutalis2009} \cite{Kottas2012} \cite{Knight2014}. While these results have been found relevant in control problems, their usability in other scenarios is less evident. Being more explicit, if the FCM converges to a unique fixed-point attractor, then the model will produce the same state vector regardless of the initial activation vector. N{\'a}poles et al. \cite{Napoles2017a} concluded that reaching a suitable tradeoff between convergence and accuracy is not always possible, while Concepci{\'o}n et al. \cite{Napoles2018a} analytically proved that the feasible state space of a sigmoid FCM will always shrink after each iteration with no guarantee of convergence to a fixed-point attractor. 

The aforementioned issues lead to the hypothesis that we can obtain lower simulation errors with an FCM-like neural model by 1) relaxing the stringent causality assumption where $w_{ij} \in [-1,1]$ and 2) limiting the number of iteration step when performing the reasoning process.        

\section{Short-term Cognitive Networks}
\label{sec:stcn}

In this section, we present a neural system baptized as \emph{Short-term Cognitive Networks} (STCNs) that attempts addressing the main drawbacks discussed above. More explicitly, we intend to develop a flexible neural model devoted to \emph{scenario simulation} where the domain experts have the responsibility to design the network architecture, while the learning algorithm is dedicated to reducing the global simulation error without modifying the expert knowledge. In order to make a clear distinction between these contributions, the present section focuses on the network architecture and reasoning, whereas the next section elaborates on the supervised learning algorithm.

\subsection{Architecture and Reasoning}
\label{sec:stcn:reasoning}

In a simulation context, an STCN can be defined as a neural network where each problem variable is denoted by a neural processing entity, whereas weights define the relation between variables. Hidden neurons are not allowed as the expert could not establish a comprehensible weight matrix due to the lack of semantics attached to these entities.

In the FCM approach, the $w_{ij}$ weight defines the type of causality between $C_i$ and $C_j$, and to what extent the $C_i$ event causes the $C_j$ event. But can we design a meaningful FCM-like reasoning model if we remove the causality assumption? We can definitely do so, although the weights interpretation will change. In the proposed STCN approach, $w_{ji} \in \mathbb{R}$ denotes the rate of change on the conditional mean of $C_i$ with respect to $C_j$, assuming that the other neurons impacting $C_i$ are fixed. Observe that this is similar to interpreting the coefficients in a multiple regression equation.

Another important component of any neural network refers to the mechanism used to propagate the initial information. Equation \ref{eq:shortmem1} and \ref{eq:shortmem2} display the neural reasoning rule attached to the proposed cognitive network,

\begin{equation}
\label{eq:shortmem1}
	A_{i}^{(t+1)}(k)=f_i^{(t+1)}\left(\sum_{j=1}^{M} w_{ji} \Psi_{j}^{(t)}(k) \right)
\end{equation}

\noindent where

\begin{equation}
\label{eq:shortmem2}
\Psi_j^{(t)}(k)= \left\{
\begin{array}{ll}
      f_j^{(t)}\left(\sum_{l=1}^{M} w_{lj} A_l^{(0)}(k) \right) & t>0 \\
      A_j^{(0)}(k) & t=0\\
\end{array} 
\right. 
\end{equation}

\noindent represents \emph{short-term evidence} attached to the $i$th neural entity in the current iteration $t$ such that $t = \{1,2,\ldots,T\}$, using the $k$th example as the excitation vector. Equation \ref{eq:transfer} formalizes the generalized sigmoid function used to squash the neuron's activation within the $[0,1]$ interval, 

\begin{equation}
\label{eq:transfer}
	f_i^{(t)}(x)=\frac{1}{(1+q_i^{(t)}e^{-\lambda_i^{(t)}(x-h_i^{(t)})})^{1/v_i^{(t)}}}
\end{equation}

\noindent where $q_i > 0$, $\lambda_i > 0$, $h_i \in {\Bbb R}$ and $v_i > 0$ are parameters that can be used to compensate the lack of flexibility imposed by the fact that weights cannot be modified, as they were fixed by the expert during the modeling stage.

\begin{figure}[!ht]
\begin{center}
\includegraphics[height=6.4cm]{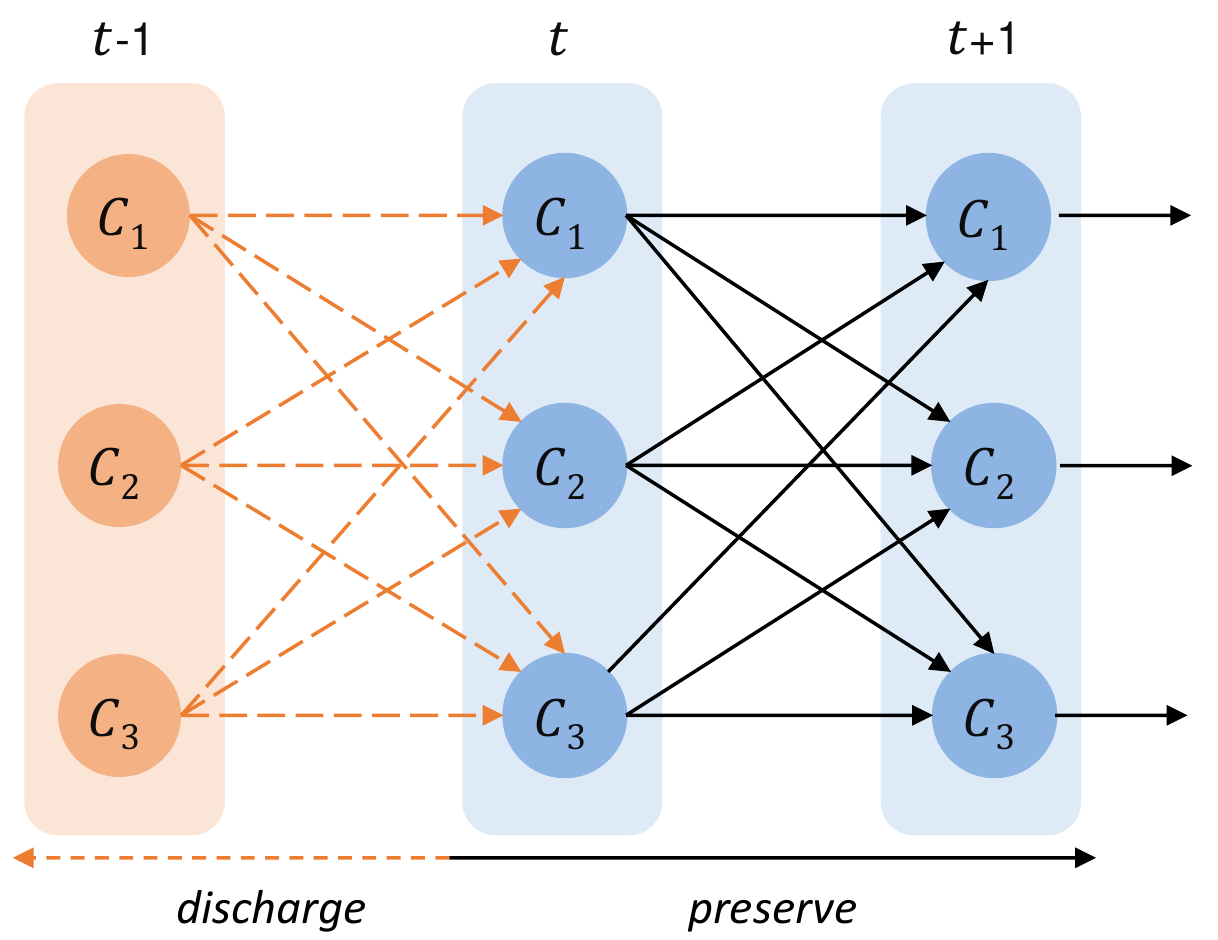} \\
\caption{Forgetting mechanism in the STCN model.}
\label{fig:forg}
\end{center}
\end{figure}

\begin{figure}[!ht]
\begin{center}
\includegraphics[height=6.4cm]{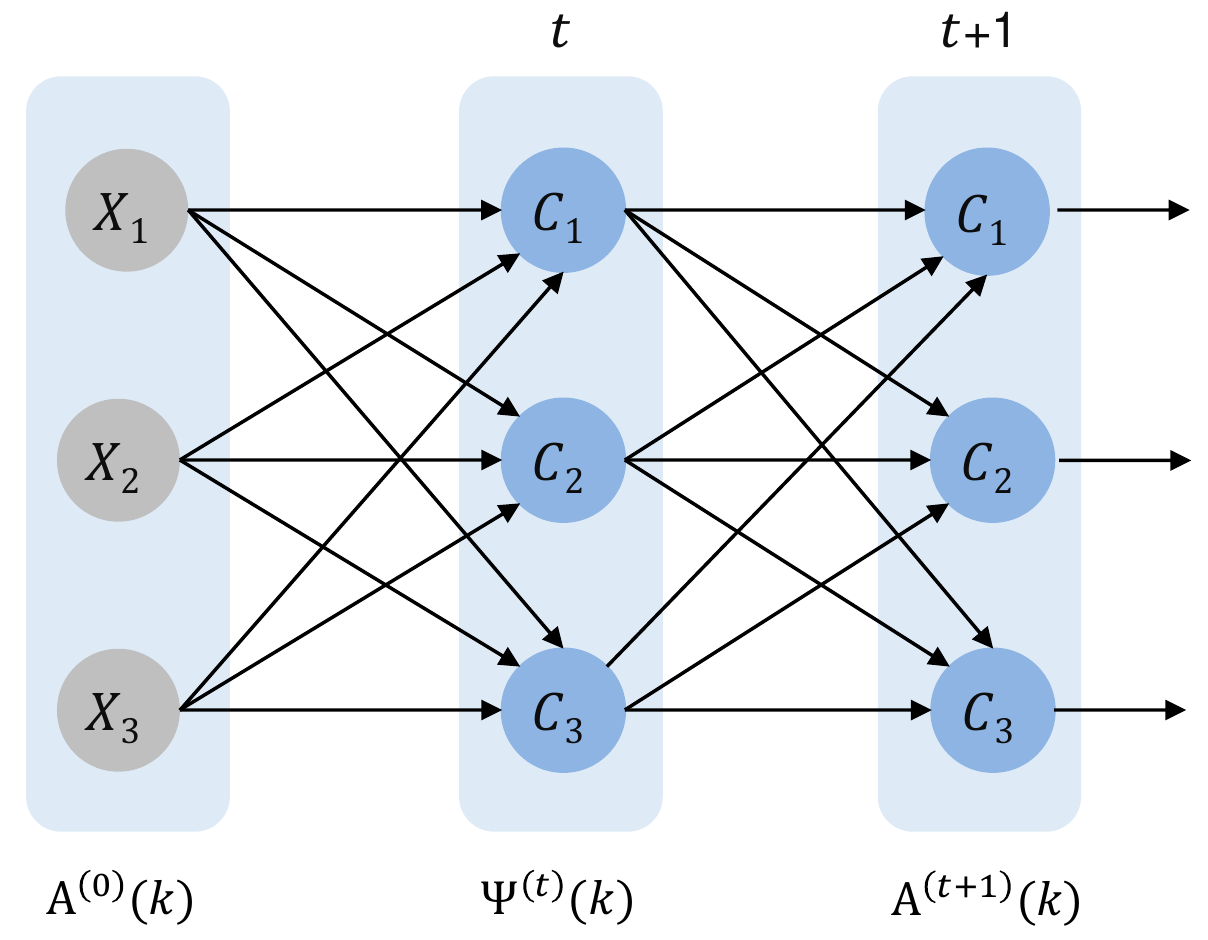} \\
\caption{Inference propagation in the STCN model.}
\label{fig:prop}
\end{center}
\end{figure}

The reasoning process of STCNs can be entirely described by two mechanisms: the forgetting step and the information propagation. During the first step (see Figure \ref{fig:forg}) the network discharges the evidence accumulated until the current iteration; only the sigmoid parameters associated with each neuron are preserved. The second step (see Figure \ref{fig:prop}) is concerned with computing the short-term evidence $\Psi_j^{(t)}(k)$ by multiplying the initial activation vector with the fixed weight matrix. As a final step, the evidence collected from neurons in the iteration $t$ is propagated to the following iteration.

In the following sub-section, we discuss how the proposed iterative neural architecture helps increase the model's accuracy when performing prediction tasks.

\subsection{Dynamical Properties}
\label{sec:stcn:dynamics}

When analyzing the way in which STCNs operate, the issues related with time series forecasting ring a bell. However, it is convenient to remark that STCN iterations and time steps are not necessarily equivalent. This implies that the our neural system can be used in presence of static data, so the subsequent iterations can be seen as a \emph{regularizer}.

As illustrated in Figures \ref{fig:forg} and \ref{fig:prop}, in each STCN  iteration the model can be effectively unfolded into a three-layer feed-forward neural network. This process can be done by mapping each iteration $t$ with a layer in the multilayer network where $w_{ji}^{(t)}=w_{ji}^{(t+1)},\forall t$. Due to the successive discharging of the old evidence it might be claimed that the system can be described by the last neural block. However, we should take into account that the forgetting operator only discharges the first abstract layer of the current neural block. 

Therefore, the STCN model does involve an implicit long-term recurrence such that the knowledge preserved from the previous iteration is stored within the neurons. This happens because the data used to adjust the sigmoid parameters in the $t+1$ iteration (during the learning phase) is obtained after transforming the original data with the  previously computed parameters. This is similar to the regularization based on data transformation \cite{Goodfellow2016} used in deep learning models.

In the STCN algorithm, the updating rule finishes once the network reaches a fixed number of iterations. If the iterations are not explicitly defined, and due to the fact that there is no a way to compute the real simulation error when exploiting the model for a unseen pattern, the maximum number of iterations will be determined during the training phase. Observe that talking about fixed-point attractors may not be realistic due to the frequent disruptions caused by the short-term reasoning memory. In spite of that, an STCN may reach a stationary state with respect to the simulation error, which does not necessarily reflect the hidden system semantics.

\section{Nonsynaptic Learning}
\label{sec:learning}

Nonsynaptic learning was first introduced and applied by N\'{a}poles et al. \cite{Napoles2016a} to improve the convergence of FCM-based models equipped with sigmoid transfer functions. In the STCN context, the nonsynaptic learning seems a suitable approach to reduce the simulation error without the need of altering the knowledge stored into the synaptic connections. This can be implemented by computing the sigmoid function parameters $q_i > 0$, $\lambda_i > 0$, $h_i \in {\Bbb R}$ and $v_i > 0$ associated with the $i$th transfer function in each iteration. 

The nonsynaptic learning procedure relies on the gradient descent method \cite{Rumelhart1986} \cite{Bengio2012}. The first step towards formalizing the learning rule is concerned with the definition of a differentiable error function for \emph{each} iteration, as we assume that the network will forget the old information due to the short-term memory. This implies that we will perform a separate learning process for iteration in an incremental fashion where only the previous sigmoid parameters are exploited, thus reducing the algorithm's complexity. Equation \ref{eq:error} shows how to compute the error of the $i$th neuron in the $t$ iteration,

\begin{equation}
\label{eq:error}
E_i^{(t)} = \sum_{k=1}^K \left(\frac{1}{\left(1+q_i^{(t)} e^{-\lambda_i^{(t)}\left(\hat{A}_{ik}^{(t)}-h_i^{(t)} \right)}\right)^{\frac{1}{v_i^{(t)}}}}-Y(k)\right)^2
\end{equation}

\noindent where $\hat{A}_{ik}^{(t)}=\sum_{j=1}^M w_{ji} \Psi_{j}^{(t-1)}(k)$ is the raw value for the $i$th neuron before being transformed, and $K$ is the number of training instances used as simulation examples. We can define a training example as a $|\mathcal{C}|$-dimensional vector comprising the activation values of all variables, where any variable in the model can be predicted from the remaining ones. 

Before applying the nonsynaptic learning method, we first transform the data into a sigmoid space since our model lacks a formal output layer with linear units, as typically occurs in multilayer perceptron networks. 

Once the sigmoid transformation step is done, we iteratively improve an initial solution until a maximum number of epochs is reached. Let us assume that the candidate solutions have the form $X=[ \lambda_i^{(t)},$ $ h_i^{(t)}, $ $q_i^{(t)}, $ $v_i^{(t)} ]$ then the gradient-based updating rule can be formalized as follows:

\begin{equation}
\label{eq:gradient}
X^{(l+1)} = X^{(l)} - Z^{(l+1)}
\end{equation} 
\begin{equation}
\label{eq:momentum}
Z^{(l+1)} = \beta Z^{(l)} + \eta \nabla E^{(t)}
\end{equation}

\noindent where $\beta \geq 0$ is the momentum, $\eta>0$ is the learning rate, while  $\nabla E^{(t)}(C_i)= \left[ \frac{\partial_E}{\partial_{\lambda_i^{(t)}}}, \frac{\partial_E}{\partial_{h_i^{(t)}}},\frac{\partial_E}{\partial_{q_i^{(t)}}}, \frac{\partial_E}{\partial_{v_i^{(t)}}} \right]$, such that

\begin{equation*}
\label{eq:der1}
\frac{\partial_E}{\partial_{\lambda_i^{(t)}}} = \sum_{k=1}^K \frac{ \gamma (\hat{A}_{ik}^{(t)}-h_i^{(t)}) (q_i^{(t)} \vartheta^{(-1-\frac{1}{v_i^{(t)}})}) (\vartheta^{-\frac{1}{v_i^{(t)}}}-Y(k))}{v_i^{(t)}},
\end{equation*}

\begin{equation*}
\label{eq:der2}
\frac{\partial_E}{\partial_{h_i^{(t)}}} =\sum_{k=1}^K -\frac{ \gamma (q_i^{(t)} \vartheta^{(-1-\frac{1}{v_i^{(t)}})}) (\vartheta^{-\frac{1}{v_i^{(t)}}}-Y(k)) \lambda_i^{(t)}}{v_i^{(t)}},
\end{equation*}

\begin{equation*}
\label{eq:der4}
\frac{\partial_E}{\partial_{q_i^{(t)}}} = \sum_{k=1}^K -\frac{ \gamma (\vartheta^{(-1-\frac{1}{v_i^{(t)}})}) (\vartheta^{-\frac{1}{v_i^{(t)}}}-Y(k))}{v_i^{(t)}},
\end{equation*}

\begin{equation*}
\label{eq:der3}
\frac{\partial_E}{\partial_{v_i^{(t)}}} =\sum_{k=1}^K \frac{ (2 \vartheta \ln (\vartheta)^{-\frac{1}{v_i^{(t)}}}) ( \vartheta^{-\frac{1}{v_i^{(t)}}}-Y(k) )}{v_i^{(t)}v_i^{(t)}},
\end{equation*}

\noindent where $\gamma$ and $\vartheta$ are defined as follows:

\begin{equation*}
\gamma = 2e^{-(\hat{A}_{ik}^{(t)}-h_i^{(t)}) \lambda_i^{(t)}}
\end{equation*}
\begin{equation*}
\vartheta = 1+q_i^{(t)} e^{-\left(\hat{A}_{ik}^{(t)}-h_i^{(t)}\right) \lambda_i^{(t)}}.
\end{equation*}

Equation \ref{eq:inverse} formalizes the inverse sigmoid function, which is employed to return the forecasted values back to the original domain. This function is not defined when $- \frac{1-y^{-v_i}}{q_i}=0$, but this scenario is not possible because $q_i>0$, $v_i>0$ and $y \neq 1$. Due to the fact that this function may produce values that grow towards infinity, we confine the real values of each variable $X_i$ to the $[L_i,U_i]$ interval, where $L_i$ and $U_i$ represent the minimum and maximum values observed in the dataset for the $i$th problem variable, respectively.

\begin{equation}
\label{eq:inverse}
f_i^{-1}(y)=\frac{- \ln \left(- \frac{1-y^{-v_i}}{q_i} \right) + h_i \lambda_i}{\lambda_i}
\end{equation}

This learning method is applied sequentially to each STCN iteration since the parameters estimated in the current iteration will be adopted to compute the following short-term evidence. Each learning process will stop either when a maximal number of epochs is reached or when the variations on the parameters from an iteration to another are infinitesimal. It can be proved that, if the latter situation comes to light, then the STCN model will enter into a stationary state that will lead to infinitesimal or no reduction in the simulation error.

\emph{Proof}. Let us assume that an implicit STCN reached a good enough local optimum in the $t+1$ iteration step, then the global simulation error for the network is:

\begin{equation*}
E^{(t+1)} = \sum_{i=1}^M \sum_{k=1}^K \left(f_i^{(t+1)}\left(\sum_{j=1}^{M} w_{ji} \Psi_{j}^{(t)}(k) \right) - Y_i(k) \right)^2
\end{equation*}

\noindent where $Y_i(k)$ represents the real (expected) value associated with the $C_i$ neuron for the $k$th simulation example, $M$ denotes the number of neural processing entities, while the short-term evidence is $\Psi_{j}^{(t)}(k)=f_j^{(t)}\left(\sum_{l=1}^{M} w_{lj} A_l^{(0)}(k) \right)$.

If $\Delta E^{(t+1)} = \left( E^{(t+1)} - E^{(t+2)} \right)^2 < \xi_1$, $t+2 \leq T$, for a small enough $\xi_1>0$, then two scenarios are possible:

\vspace{3mm}

\begin{itemize}
 \item {$F_1 = \{f_i^{(t+1)}\}_{i=1}^M$ and $F_2 = \{f_i^{(t+2)}\}_{i=1}^M$ represent two different local optima. Consequently,
\begin{equation*}
\exists C_i \in \mathcal{C} \mid \int_{-M}^{+M} \left( f_i^{(t+1)}(x) - f_i^{(t+2)}(x) \right)^2 dx > \xi_2.
\end{equation*}
}
 
\item {$F_1 = \{f_i^{(t+1)}\}_{i=1}^M$ and $F_2 = \{f_i^{(t+2)}\}_{i=1}^M$ represent the same local optimum. Consequently,
\begin{equation*}
\nexists C_i \in \mathcal{C} \mid \int_{-M}^{+M} \left( f_i^{(t+1)}(x) - f_i^{(t+2)}(x) \right)^2 dx > \xi_2.
\end{equation*}
}
\end{itemize}

If the first scenario comes to light, then we are in presence of a multimodal space since the families of sigmoid functions $F_1$ and $F_2$ lead to equally accurate solutions, but there exist at least a pair of functions $f_i^{(t+1)}(x) \in F_1$ and $f_i^{(t+2)}(x) \in F_2$ that differ in their shapes. This implies that the network could still have a chance to reach a better solution, thus we should continue iterating. In the second case, the shapes of all pairs of functions $f_i^{(t+1)}(x) \in F_1$ and $f_i^{(t+2)}(x) \in F_2$ are significantly similar. For a small enough $\xi_2$ and given the fact that STCNs employ a short-term reasoning process on which weights do not change from an iteration to the following, we can claim that $\Psi_{j}^{(t+1)}(k) \approx \Psi_{j}^{(t+2)}(k), \forall j,k$ because  

\begin{equation*}
f_j^{(t+1)}\left(\sum_{l=1}^{M} w_{lj} A_l^{(0)}(k) \right) \approx f_j^{(t+2)}\left(\sum_{l=1}^{M} w_{lj} A_l^{(0)}(k) \right).
\end{equation*}
\\
On the other hand, since our learning method is deterministic, it is reasonable to expect the same behavior for the functions in $\{f_i^{(t+3)}\}_{i=1}^M, \ldots, \{f_i^{(T-1)}\}_{i=1}^M$, therefore

\begin{equation*}
\int_{-M}^{+M} \left( f_i^{(p)}(x) - f_i^{(p+1)}(x) \right)^2 dx \approx 0
\end{equation*}

such that $p=\{t+1, t+2, \ldots, T-1\}$. Given the fact that the sigmoid activator is a continuous, non-noisy function we can claim that $\Delta E^{(p)} \approx 0$. This suggests that the STCN reached a stationary state on which further learning iterations will not significantly reduce the simulation error. \qedsymbol

\section{Numerical Simulations}
\label{sec:sim}

In this section, we evaluate the performance of STCNs in simulation scenarios where the goal is to forecast the value of any variable from the remaining ones.

\subsection{Proof of concept}
\label{sec:sim:example}

In our first experiment, we illustrate how STCNs operate for the Iris problem. This dataset contains 3 decision classes of 50 instances each, where classes refer to the type of plant. One class is linearly separable from the others, while the other two are not linearly separable from each other. As we are not interested in solving the classification problem, we remove the decision class during the preprocessing stage. 

The first step in modeling this problem is to represent each feature with a sigmoid neuron. As an alternative to the lack of expert knowledge for this problem, we can model the relation between neurons $C_j$ and $C_i$ as a linear regression equation $A_j=w_{ji}A_i + r_j$. Equations \ref{eq:weight} and \ref{eq:residual} show how to compute such coefficients from the dataset,

\begin{equation}
\label{eq:weight}
w_{ji} = \frac{K \sum_{k} x_k(j) x_k(i) - \sum_{k} x_k(j) \sum_{k} x_k(i)}{K(\sum_{k} x_k(j)^2)-(\sum_{k} x_k(j))^2}
\end{equation}

\begin{equation}
\label{eq:residual}
r_j = \frac{\sum_{k} x_k(i)^2 \sum_{k} x_k(j) - \sum_{k} x_k(i) x_k(j) \sum_{k} x_k(i)}{K(\sum_{k} x_k(i)^2)-(\sum_{k} x_k(i))^2}
\end{equation}

\noindent where $K$ denotes the number of instances in the training set, whereas $x_k(i)$ represents the value of the $i$-th variable as coded in the $k$-th training example. It should be stated that the $r_j$ coefficients are used to initialize the offset parameter of the $i$th neuron as $h_i^{(0)}=\min\{1,\sum_{j=1}^{M}r_j\}$, which simulates the bias in a McCulloch-Pitts neuron \cite{McCulloch1988}. This initialization strategy may help reach better results with less effort when training neural system. The remaining sigmoid function parameters are internalized using a random approach 

\begin{figure}[!ht]
\begin{center}
\includegraphics[height=8cm]{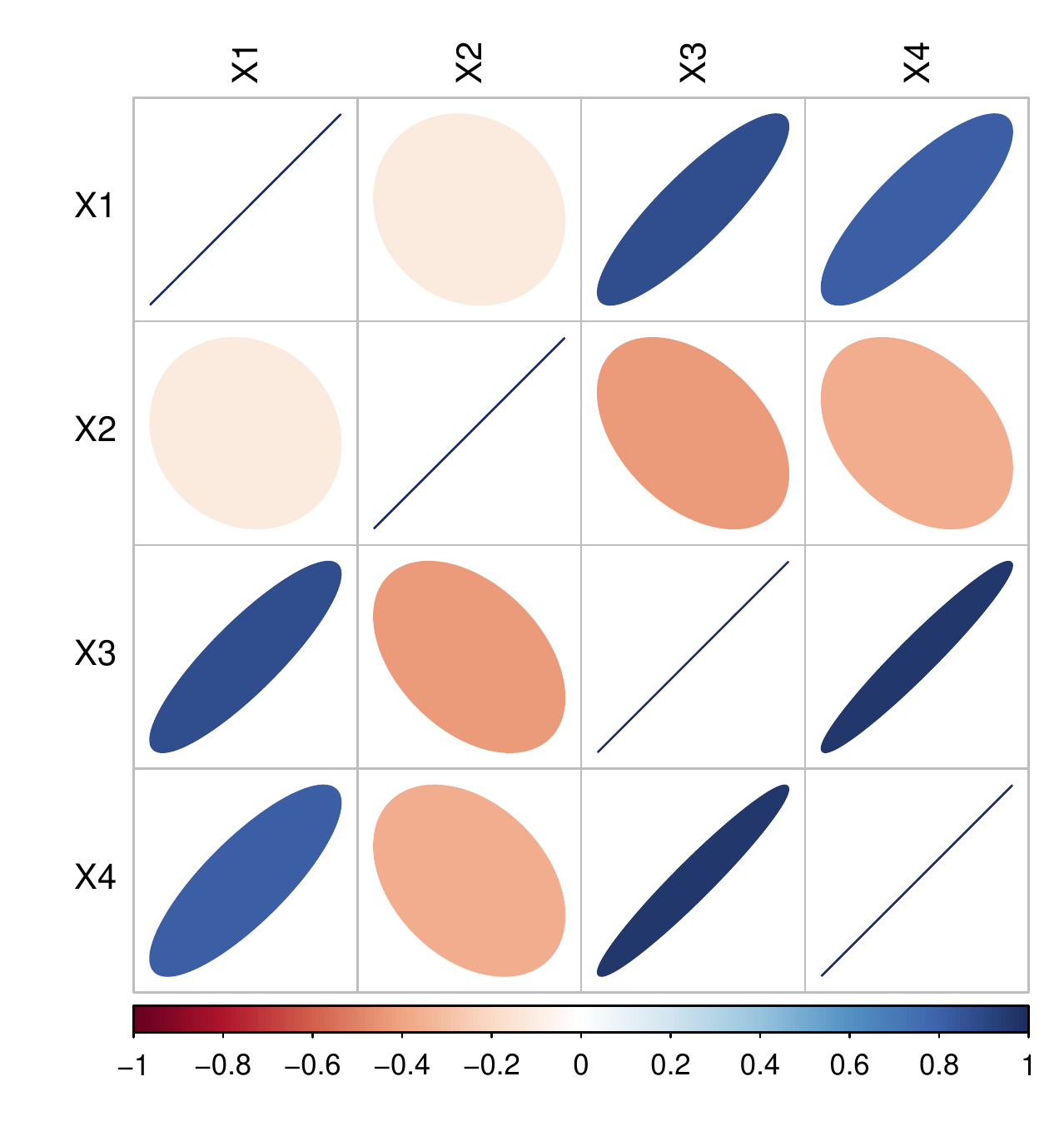} \\
\caption{Pearson correlation between Iris variables.}
\label{fig:corr-iris}
\end{center}
\end{figure}

Figure \ref{fig:corr-iris} portrays the Pearson correlation between variables describing the Iris dataset. This matrix provides a clear picture of the strength of each synaptic connection on the resulting nonlinear neural system. As a next step, we fine-tune the model performance by using the gradient descent learning algorithm discussed in the previous section.

In all simulations conducted in this paper, the learning rate is set to 0.001, the momentum is set to 0.85, while the number of epochs for each training process is set to 500. Since our neural system resembles a multiple regression, we adopted several state-of-the-art models for comparison purposes, which are implemented in the R programming language. The selected algorithms include: linear regression (LREG) from the \texttt{nnet} package, support vector regression (SVM) from the \texttt{e1071} package, $k$-nearest neighbors ($k$NN) from the \texttt{class} package, random forest (RF) from the \texttt{randomForest} package and multilayer perceptron (MLP) from the \texttt{monmlp} package. It is fair to mention that we have relied on the default parameters attached to the selected methods, therefore no algorithm performs parameter tuning. Although a proper parametric setting regularly increases the algorithm's performance over multiple data sources, a robust model should be capable of producing good results even when its parameters might not have fully been optimized for a specific problem.

\begin{figure}[!ht]
\includegraphics[height=6.0cm]{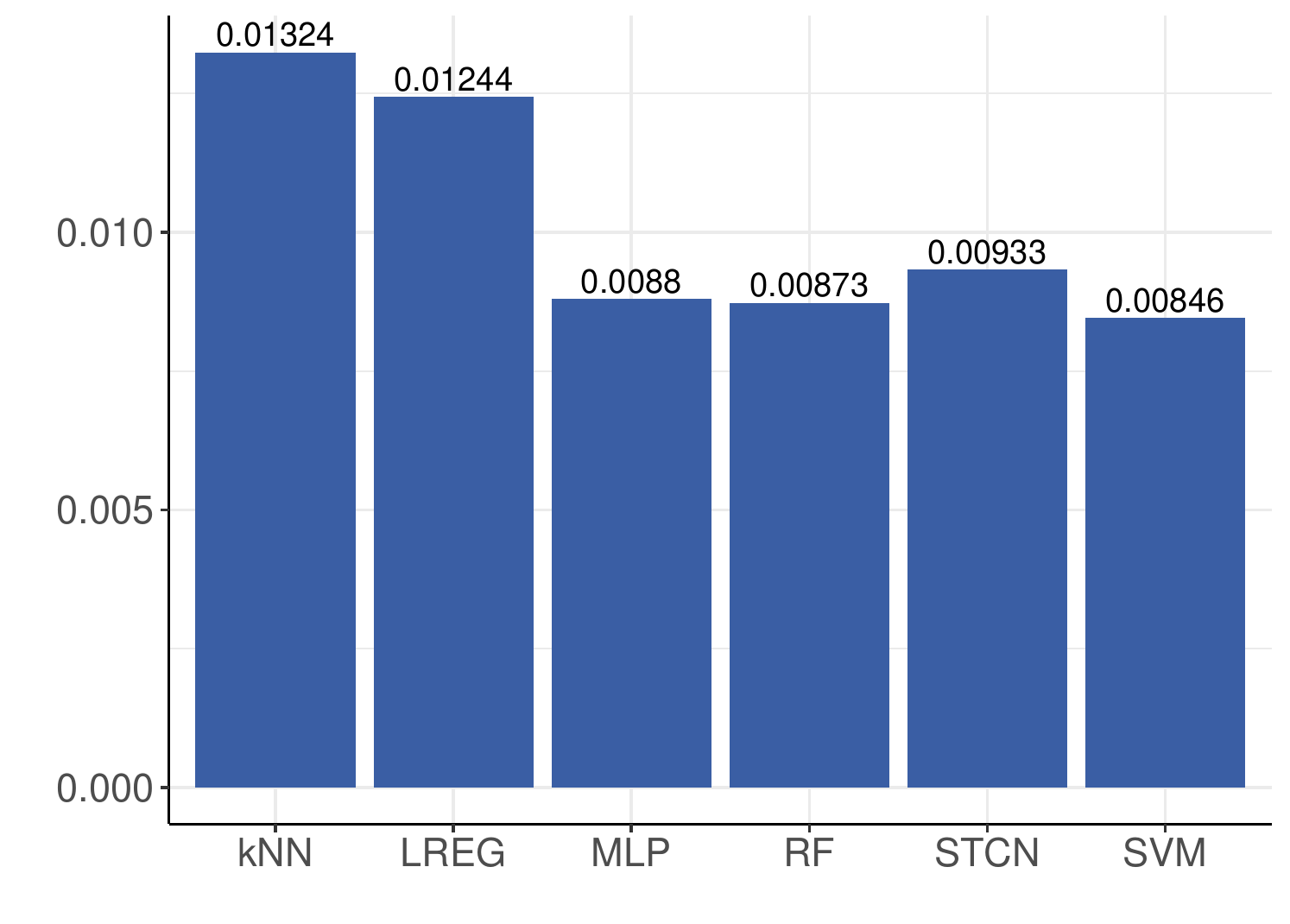} \\
\caption{MSE achieved by each regression model for the Iris dataset using a 10-fold cross-validation.}
\label{fig:cmp1}
\end{figure}

Figure \ref{fig:cmp1} depicts the Mean Squared Error (MSE) obtained by each algorithm, after performing a 10-fold cross-validation process. In the case of the state-of-the-art regression methods, we report the MSE across several independent models (i.e., one per each dependent variable) since these algorithms allow to predict a single variable at each time. Hence, each variable in the dataset will be used one time as the dependent variable, and $N$-1 times as an independent variable. The simulations show that STCNs are able to produce competitive approximations without the need of building a new regression model for each dependent variable to be predicted. 

\begin{figure}[!ht]
\begin{center}
\includegraphics[height=5cm]{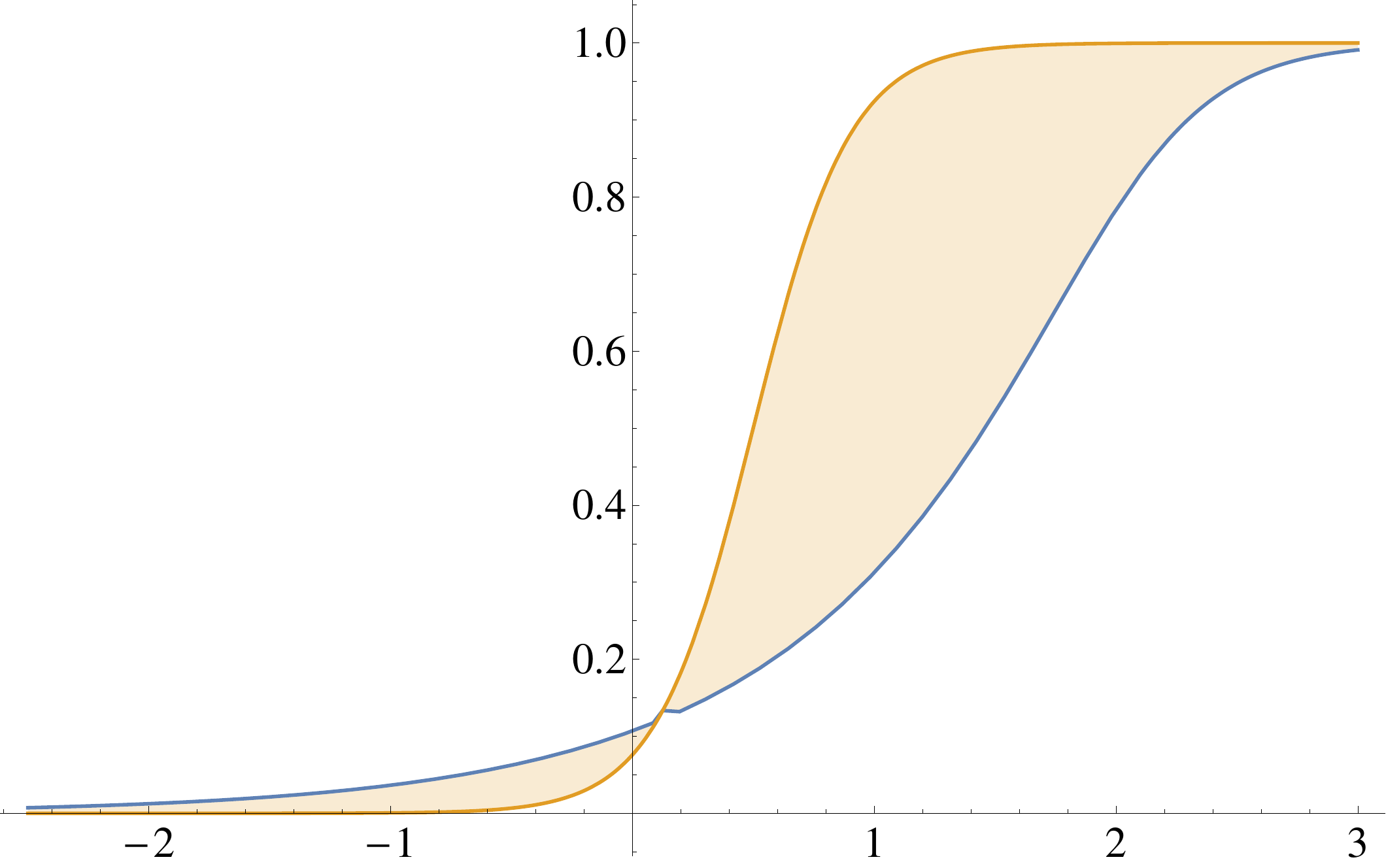} \\
\caption{Sigmoid function adjustment for the X1 neuron at the beginning of the reasoning process.}
\label{fig:sigmoid1}
\end{center}
\end{figure}

\begin{figure}[!ht]
\begin{center}
\includegraphics[height=5cm]{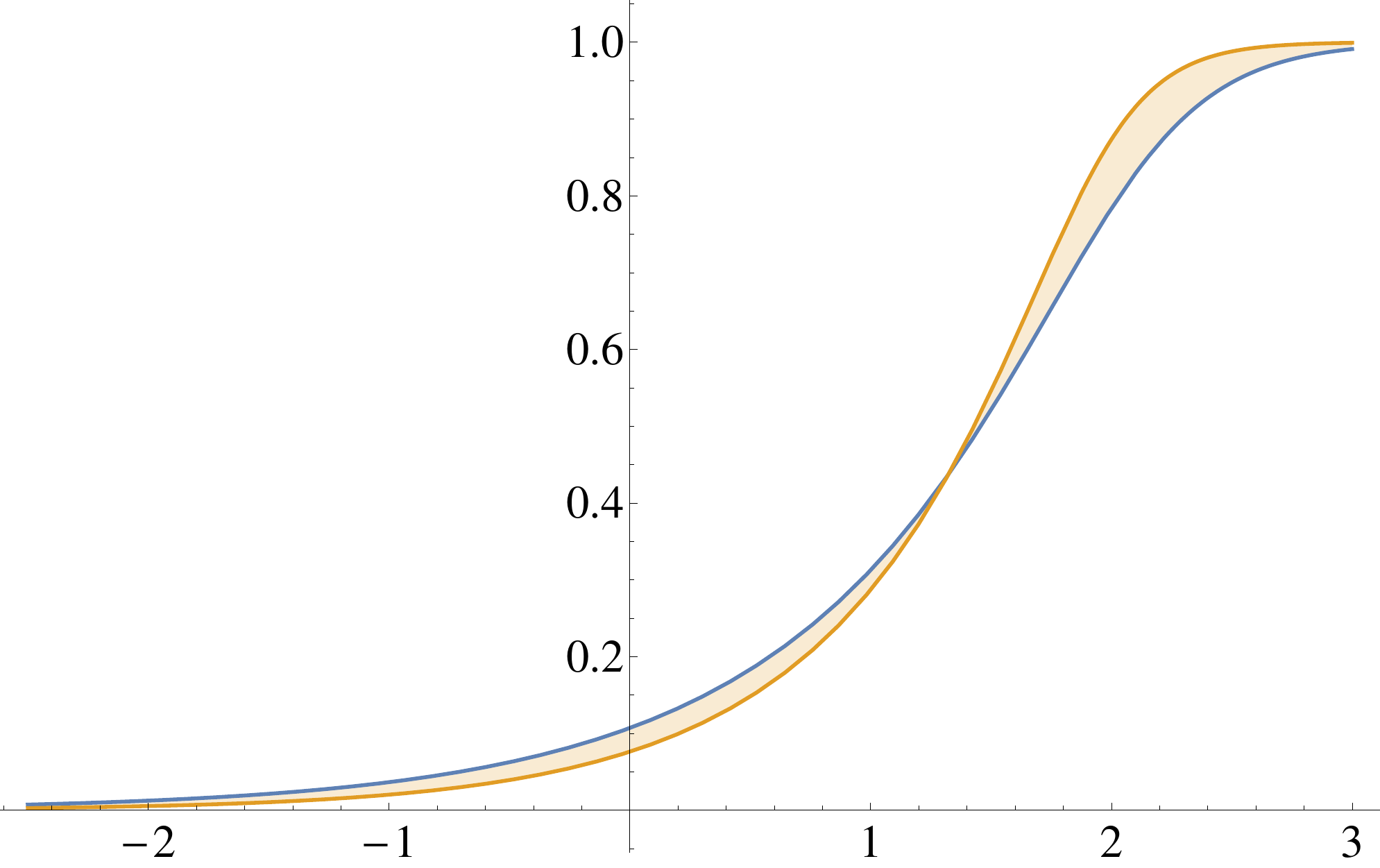} \\
\caption{Sigmoid function adjustment for the X1 neuron towards the end of the reasoning process.}
\label{fig:sigmoid2}
\end{center}
\end{figure}

Figure \ref{fig:sigmoid1} and \ref{fig:sigmoid2} display, as an example, the sigmoid function adjustment for the the neuron denoting the $X1$ variable (i.e., sepal length). The orange line represents the sigmoid function in the previous iteration step, while the blue one indicates the adjusted function. In this simulation, the initial solution was set as follows: $\lambda_1=5$, $h_1=0.5$, $q_1=1$ and $v_1=1$. After a few iterations the learning algorithm was able to decrease the simulation error from 0.3789 to 0.0007, without changing the weight matrix configuration.

\begin{figure}[!ht]
\begin{center}
\includegraphics[height=5cm]{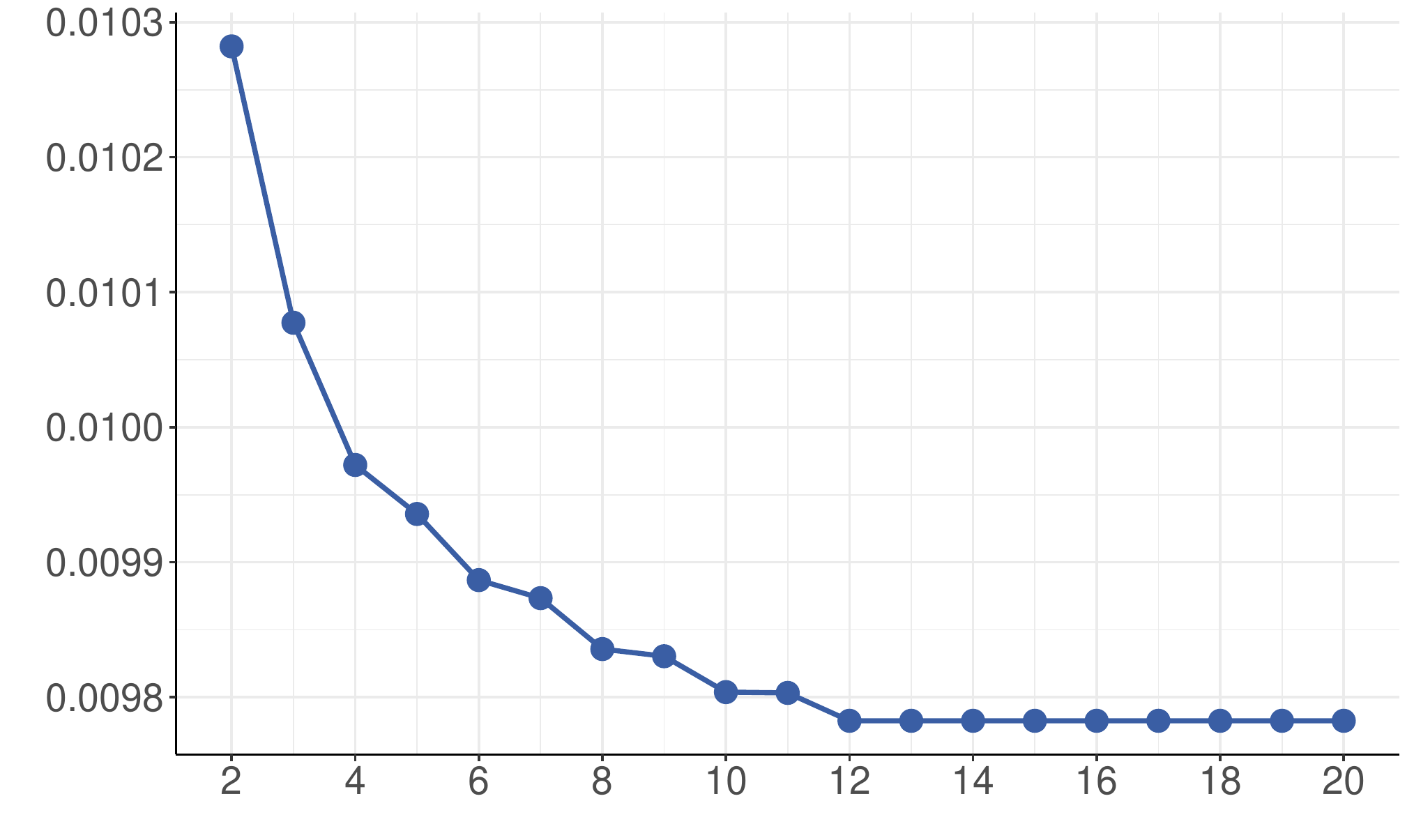} \\
\caption{Global simulation error at each layer.}
\label{fig:error-iris}
\end{center}
\end{figure}

Figure \ref{fig:error-iris} portrays the global simulation error attained by the network where all available data is used for both training and testing, so the focus is not the algorithm's generalization ability. Instead, the goal is to illustrate how the error decreases from an STCN iteration to the following, until a local optimum is discovered. Towards the end, the system becomes stationary since there is no reason to update the shape of the sigmoid function attached to each neural entity.

In general, the preliminary results confirm that the proposed STCN algorithm (in conjunction with the nonsynaptic learning approach) is suitable to perform simulations on the basis of a previously defined weight matrix. It is worth mentioning that our model heavily depends on that knowledge, and thus we cannot expect high-quality predictions from an STCN network comprising a poorly defined weight matrix.

\subsection{Extended experiments: multiple regression}
\label{sec:sim:ext}

Aiming at expanding our analysis with more challenging scenarios, we have selected 35 datasets from the study reported in \cite{Napoles2017c}. Table \ref{table:datasets} outlines the key features of such datasets where the number of attributes ranges from 3 to 22, the number of decision classes from 2 to 8, whereas the number of instances goes from 106 to 625. Similarly to the previous experiment, we remove the decision class since our goal is to predict the value of numerical attributes. Observe that our experiments use rather small datasets for the following reasons: (i) to keep the simulation time low, and (ii) to better resemble the real-world simulation scenarios on which experts are able to provide a limited number of training examples.

\begin{table}[!ht]
\centering
\caption{Datasets used during simulations.}
\label{table:datasets}
\begin{tabular}{|llcccc|}
\hline
Id & Dataset & Instances & Attributes & Classes & Noisy \\
\hline
1 & acute-inflammation & 120 & 6 & 2 & no \\
2 & acute-nephritis & 120 & 6 & 2 & no \\
3 & appendicitis & 106 & 7 & 2 & no \\
4 & balance-noise & 625 & 4 & 3 & yes \\
5 & balance-scale & 625 & 4 & 3 & no \\
6 & blood & 748 & 4 & 2 & no \\
7 & echocardiogram & 131 & 11 & 2 & no \\
8 & ecoli & 336 & 7 & 8 & no \\
9 & glass & 214 & 9 & 6 & no \\
10 & glass-10an-nn & 214 & 9 & 6 & yes \\ 
11 & glass-20an-nn & 214 & 9 & 6 & yes \\
12 & glass-5an-nn & 214 & 9 & 6 & yes \\
13 & haberman & 306 & 3 & 2 & no \\
14 & hayes-roth & 160 & 4 & 3 & no \\  
15 & heart-5an-nn & 270 & 13 & 2 & yes \\
16 & heart-statlog & 270 & 13 & 2 & no \\
17 & iris & 150 & 4 & 3 & no \\
18 & iris-10an-nn & 150 & 4 & 3 & yes \\
19 & iris-20an-nn & 150 & 4 & 3 & yes \\
20 & iris-5an-nn & 150 & 4 & 3 & yes \\
21 & liver-disorders & 345 & 6 & 2 & no \\ 
22 & monk-2 & 432 & 6 & 2 & no \\
23 & new-thyroid & 215 & 5 & 2 & no \\
24 & parkinsons & 195 & 22 & 2 & no \\
25 & pima & 768 & 8 & 2 & no \\
26 & pima-10an-nn & 768 & 8 & 2 & yes \\
27 & pima-20an-nn & 768 & 8 & 2 & yes \\
28 & pima-5an-nn & 768 & 8 & 2 & yes \\
29 & planning & 182 & 12 & 2 & no \\
30 & saheart & 462 & 9 & 2 & no \\
31 & tae & 151 & 5 & 3 & no \\
32 & vertebral2 & 310 & 6 & 2 & no \\
33 & vertebral3 & 310 & 6 & 3 & no \\
34 & wine & 178 & 13 & 3 & no \\
35 & wine-5an-nn & 178 & 13 & 3 & yes \\
\hline
\end{tabular}
\begin{tablenotes}
\item[a]* From the study in \cite{Napoles2017c}, we have discarded the problems with high number of attributes or instances. Additionally, we have omitted the datasets comprising nominal attributes or those involving class imbalance since they are described by the same attributes, thus resulting in 35 problems.
\end{tablenotes}
\end{table}

\begin{figure}[!ht]
\includegraphics[height=6.0cm]{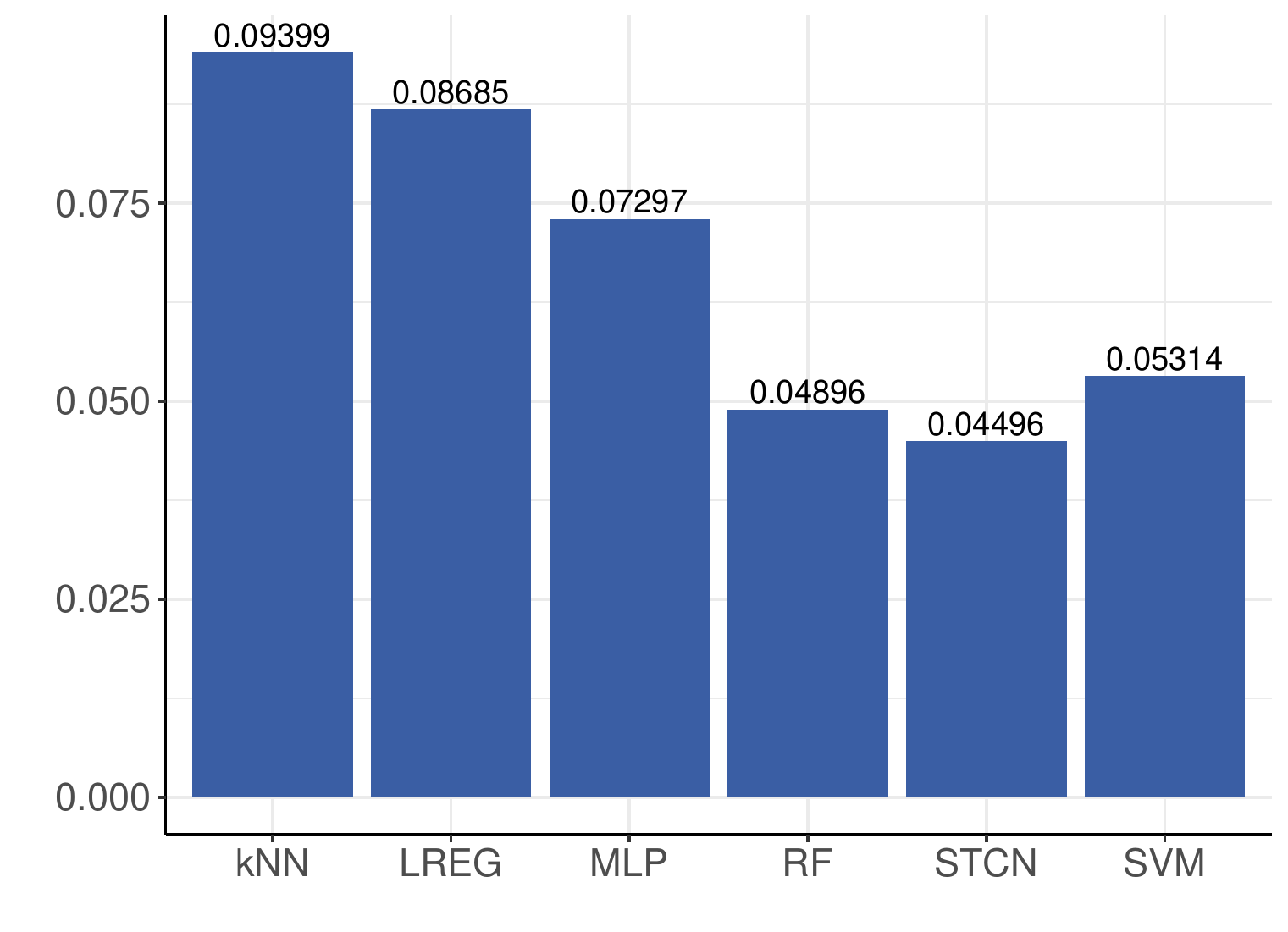} \\
\caption{Average MSE attained by each regression model across datasets using a 10-fold cross-validation.}
\label{fig:summary}
\end{figure}

Table \ref{table:benchmark} shows the MSE achieved by each algorithm across selected datasets, which are further summarized in Figure \ref{fig:summary}. From these results we can conclude that the proposed model is the best-performing technique followed by RF, while SVM is ranked at the third position. It is remarkable the superiority of our neural approach with respect to these well-established algorithms, even when they build an independent regression model per each dependent variable to be predicted. Or perhaps, being capable of approximating the value of multiple variables using a single model eventually becomes a key piece towards producing lower simulation errors.

\begin{table}[!ht]
\centering
\caption{MSE achieved by each algorithm.}
\label{table:benchmark}
\begin{tabular}{|lcccccc|}
\hline
  & STCN & MLP & LREG & RF & SVM & kNN \\
\hline
1 & 0.03787 & 0.07916 & 0.04026 & 0.06733 & 0.08148 & 0.06957 \\
2 & 0.04400 & 0.08382 & 0.04571 & 0.07087 & 0.08117 & 0.06729 \\
3 & 0.02044 & 0.01196 & 0.00133 & 0.01806 & 0.00790 & 0.00708 \\
4 & 0.12450 & 0.30100 & 0.13496 & 0.29450 & 0.14670 & 0.13570 \\
5 & 0.12504 & 0.30767 & 0.13387 & 0.28762 & 0.14340 & 0.13894 \\
6 & 0.05630 & 0.01513 & 0.00819 & 0.01658 & 0.00962 & 0.00944 \\
7 & 0.04917 & 0.11334 & 0.11955 & 0.08318 & 0.05374 & 0.06031 \\
8 & 0.02363 & 0.05841 & 0.04722 & 0.05698 & 0.03337 & 0.03433 \\
9 & 0.01878 & 0.01663 & 0.00689 & 0.02022 & 0.01128 & 0.01283 \\
10 & 0.02441 & 0.04765 & 0.16209 & 0.05201 & 0.02813 & 0.02945 \\
11 & 0.03647 & 0.07961 & 0.07921 & 0.07994 & 0.04557 & 0.05014 \\
12 & 0.02249 & 0.03937 & 0.08893 & 0.04100 & 0.02452 & 0.02471 \\
13 & 0.04913 & 0.09917 & 0.05285 & 0.09786 & 0.05471 & 0.05435 \\
14 & 0.11574 & 0.24857 & 0.14463 & 0.21081 & 0.12598 & 0.13820 \\
15 & 0.08662 & 0.21027 & 0.20435 & 0.16798 & 0.10611 & 0.11303 \\
16 & 0.08213 & 0.17867 & 0.17540 & 0.14922 & 0.09663 & 0.10388 \\
17 & 0.01019 & 0.01324 & 0.00880 & 0.01244 & 0.00873 & 0.00846 \\
18 & 0.02148 & 0.05321 & 0.04363 & 0.05324 & 0.03280 & 0.03189 \\
19 & 0.02955 & 0.07236 & 0.05407 & 0.06804 & 0.04248 & 0.04414 \\
20 & 0.01664 & 0.04007 & 0.03117 & 0.04204 & 0.02360 & 0.02605 \\
21 & 0.03707 & 0.03155 & 0.01733 & 0.02460 & 0.01534 & 0.01646 \\
22 & 0.19033 & 0.63718 & 0.25410 & 0.46844 & 0.21145 & 0.31691 \\
23 & 0.00590 & 0.01718 & 0.02657 & 0.01758 & 0.01029 & 0.01115 \\
24 & 0.02139 & 0.01190 & 0.01584 & 0.01829 & 0.00670 & 0.01031 \\
25 & 0.01877 & 0.03355 & 0.02287 & 0.03606 & 0.01804 & 0.01888 \\
26 & 0.03529 & 0.06430 & 0.03809 & 0.07540 & 0.03406 & 0.03608 \\
27 & 0.04886 & 0.08827 & 0.05114 & 0.10189 & 0.04735 & 0.04926 \\
28 & 0.02759 & 0.04929 & 0.03025 & 0.05818 & 0.02657 & 0.02813 \\
29 & 0.01539 & 0.02042 & 0.00225 & 0.03323 & 0.00976 & 0.00792 \\
30 & 0.04083 & 0.08351 & 0.05175 & 0.08130 & 0.04642 & 0.05310 \\
31 & 0.07517 & 0.09295 & 0.34300 & 0.12766 & 0.07266 & 0.09862 \\
32 & 0.01047 & 0.01244 & 0.00661 & 0.01220 & 0.00851 & 0.00626 \\
33 & 0.01034 & 0.01227 & 0.00624 & 0.01237 & 0.00843 & 0.00630 \\
34 & 0.01914 & 0.02831 & 0.05457 & 0.03604 & 0.01746 & 0.01755 \\
35 & 0.02249 & 0.03729 & 0.05054 & 0.04691 & 0.02293 & 0.02334 \\
\hline
\end{tabular}
\end{table}

Likewise, we can observe that our algorithm reports a stable behavior in situations where other methods report error peaks. Again, this could be a positive consequence of predicting the value of multiple dependent variables using a single simulation model. In other words, during the reasoning phase, the STCN algorithm attempts to reduce the overall network error, even when the simulation error associated with a particular neural processing entity slightly increases. 

Aiming at exploring whether the differences in algorithms' performance are statistically significant or not, we rely on the Friedman two-way analysis of variances by ranks \cite{Friedman1937a}. This test advocates for the rejection of the null hypothesis ($p$-value = 1.16576E-16 $<$ 0.05) for a confidence interval of 95\%, hence we can conclude that there are significant differences between at least two models across datasets.

Next, we perform a pairwise significance analysis using the STCN model as the control method. By doing so, we resorted to the Wilcoxon signed rank test \cite{Wilcoxon1945a} and post-hoc procedures to adjust the $p$-values instead of using mean-ranks approaches, as recently suggested by Benavoli et al. \cite{Benavoli2016a}. Table \ref{post-hoc-reg} reports the unadjusted $p$-value computed by the Wilcoxon signed rank test and the corrected $p$-values associated with each pairwise comparison. In this research, we assume that a null hypothesis can be rejected for a certain confidence level if at least one of the post-hoc procedures supports the rejection.

\begin{table}[!ht]
\centering
\caption{Pairwise analysis using the STCN algorithm as the control method (multiple regression setting).}
\label{post-hoc-reg}
\begin{tabular}{|lcccc|}
\hline
Algorithm & $p$-value & Bonferroni & Holm & Holland  \\
\hline
LREG & 7.767E-6 & 3.883E-5 & 3.883E-5 & 3.883E-5 \\
kNN  & 1.224E-5 & 6.121E-5 & 4.896E-5 & 4.896E-5 \\
MLP  & 0.001858 & 0.009290 & 0.005574 & 0.005574 \\
SVM  & 0.013389 & 0.066945 & 0.026778 & 0.026778 \\
RF   & 0.079676 & 0.398380 & 0.079676 & 0.079676 \\
\hline
\end{tabular}
\end{table}

The statistical analysis reveals that the proposed algorithm is superior in performance to LREG, kNN, MLP and SVM since both Holm and Holland suggest to reject the null hypotheses for a 95\% confidence interval. In the case of the STCN-RF pair, both Holm and Holland can support the rejection of the null hypothesis in favor of our algorithm if we adopt a 90\% confidence interval. In a nutshell, these results evidence the capability of our algorithm to produce high-quality predictions with less effort, i.e. without the need of building a separate model for each dependent variable. In spite of that, it should be highlighted that the main contribution attached to our proposal is the flexible reasoning scheme that allows including expert knowledge into the network structure.

\subsection{Extended experiments: associative memories}
\label{sec:sim:mem}

In the previous section we evaluated the prediction capability of STCNs against traditional regression models. However, our neural system goes beyond multiple regression in the sense that it can handle prediction problems involving several dependent variables simultaneously, which closely resembles the way associative memories operate \cite{Krikelis1996a}. As a representative of these networks we have selected the Hopfield model \cite{Hopfield84a} since its architecture is reasonably similar to the one discussed in this paper, i.e. each neuron has a well-defined meaning for the prediction problem under analysis.

Likewise, we compare our method against an FCM using sigmoid neurons ($\lambda=q=v=1, h=0$) where weights can be freely distributed in the $[-1,1]$ interval. One advantage of FCMs over other techniques is that they can be designed to simulate an associative memory, that is why we have reserved them for this section. Aiming at estimating the weight set, we adopt an evolutionary learning approach based on an elitist Real-Coded Genetic Algorithm \cite{Stach2005}. The required parameters are fixed as follows: the mutation probability is set to 0.1, the crossover probability is set to 0.9, the number of generations is set to 100, whereas the number of chromosomes is equal to the number of weights to be estimated.  

With respect to the data requirements for this experiment, we modify the datasets listed in Table \ref{table:benchmark} so that each record has probability 0.2 to be corrupted (i.e., replaced with zero). Embracing this transformation instead of just using datasets with missing data facilitates the evaluation because we know beforehand the values to be expected. 

\begin{figure*}[!ht]
\begin{center}
\includegraphics[height=6.4cm]{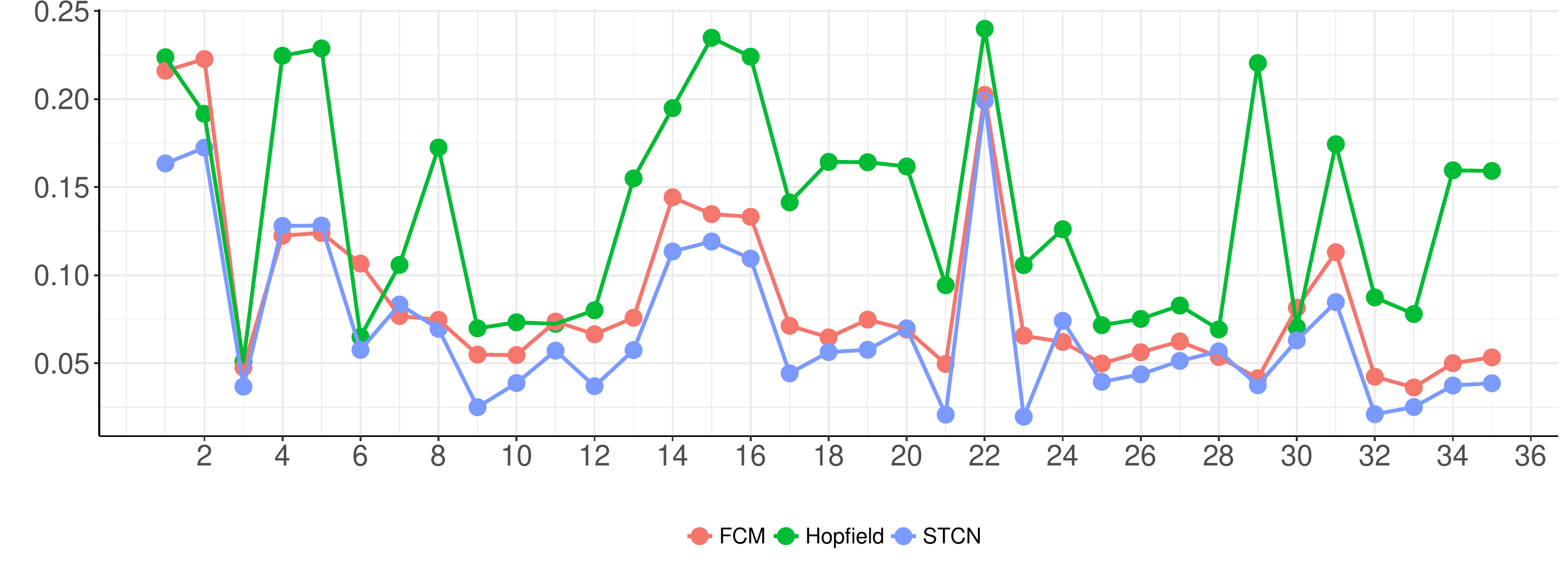} \\
\caption{MSE achieved by each associative memory across modified datasets.}
\label{fig:associative}
\end{center}
\end{figure*}

Figure \ref{fig:associative} shows the performance of each associative neural network after performing a 10-fold cross-validation, where the vertical axis indicates the MSE and the horizontal axis encodes the identifier of each dataset. The results suggest that our proposal is capable of outperforming the other algorithms, while Hopfield reports the higher prediction errors; this could be a direct result of using a Hebbian learning procedure over datasets described by rather small patterns.  

The Friedman test confirms that there are statistically significant differences in the performance of the above algorithms ($p$-value=6.2402E-12 $<$ 0.05) for a confidence interval of 95\%. Table \ref{post-hoc-reg} depicts the unadjusted $p$-value obtained with the Wilcoxon signed rank test and the adjusted $p$-values. All post-hoc procedures advice rejecting the conservative hypothesis for a significance level of 0.05, thus corroborating the superiority of the STCN algorithm in this experiment. 

\begin{table}[!ht]
\centering
\caption{Pairwise analysis using the STCN algorithm as the control method (associative memory setting).}
\label{post-hoc-ass}
\begin{tabular}{|lcccc|}
\hline
Algorithm & $p$-value & Bonferroni & Holm & Holland  \\
\hline
Hopfield & 2.477E-7 & 4.954E-7 & 4.954E-7 & 4.954E-7 \\
FCM      & 5.278E-6 & 1.055E-5 & 5.278E-6 & 5.278E-6 \\
\hline
\end{tabular}
\end{table}

Overall, the results have shown that our neural system is a convenient simulation technique in terms of simulation error. Remark that our model is perfectly suited to deal with static data since abstract layers resulting from the iterations act as a regularizer. However, the forecasting of times series with STCNs emerges as the most logical step towards exploring the potential behind our model.  

\section{Concluding Remarks}
\label{sec:conclusions}

In this research, we have presented a flexible neural system referred to as \emph{Short-term Cognitive Networks} as an alternative to classic FCMs. A key feature of the STCN model is that it allows performing simulations on the basis of previously defined knowledge structures, where weights may have a causal meaning or not. Aiming at preserving the initial knowledge, we developed a nonsynaptic learning algorithm that relies on the gradient descent to reduce the global error without altering the weight set. Furthermore, we have analytically derived a stopping criterion to prevent the learning method from iterating without decreasing the simulation error.

The statistical analysis reported that the STCN model is superior (in terms of simulation error) to most regression models adopted for comparison. Moreover, our proposal allows making predictions without the need of building a separate model for each decision variable to be forecasted. In the context of associative memories, the results support the superiority of our model over both FCMs and  Hopfield.

While our proposal is accurate in terms of simulation errors, it brings to life a sensitive problem: the short-term reasoning mechanism is not able to capture the dynamics governing the system under analysis. Therefore, another point that should be studied is how to expand the STCN memory without affecting the prediction rates, which would allow capturing the dynamic patterns attached to the system. But if an explicit long-term memory brings to life the limitations inherent to classic FCMs, then it would be a high price to be paid on behalf of patterns that experts rarely exploit in practice. 

Although this research focused on providing an accurate alternative for traditional FCMs used in simulation scenarios, the STCN model can certainly be expanded to other domains. Time series forecasting or pattern classification based on the flexible reasoning paradigm are open problems that deserve attention. Another interesting challenge that may be explored in futures studies is how to efficiently fine-tune the network parameters in presence of time-varying pieces of data. Likewise, how to handle the uncertainty introduced by experts during the modeling stage is deemed pivotal towards designing a robust, still flexible, neural reasoning model.
 
\section*{Acknowledgements}
The authors would like to thank the Business Intelligence research group of Hasselt University for the critical remarks and constructive suggestions.

\bibliographystyle{IEEEtran}
\bibliography{references}


\end{document}